\documentclass{article}

\usepackage{amsmath}
\usepackage{amsthm}
\usepackage{amssymb}
\usepackage{booktabs} 
\usepackage{caption}
\usepackage{graphicx}
\usepackage{microtype}
\usepackage{subcaption}
\usepackage{url}
\usepackage{enumitem}
\usepackage{todonotes}
\usepackage[normalem]{ulem}

\usepackage{hyperref}

\usepackage[accepted]{icml2020}

\newtheorem{prop}{Proposition}

\newcommand{\refeq}[1]{Eq.~\eqref{#1}}
\newcommand{\refsec}[1]{Sec.~\ref{#1}}
\newcommand{\reffig}[1]{Fig.~\ref{#1}}

\newcommand{\E}{\mathbb{E}}
\newcommand{\var}{\mathop{\mathbb{V}\textrm{ar}}}

\newcommand{\norm}[1]{\left\lVert#1\right\rVert}

\newcommand{\minvar}{MinVar}

\newcommand{\directloss}{Direct Loss}

\newcommand{\parameters}{\mathbf{\theta}}

\icmltitlerunning{Defense Through Diverse Directions}

\begin{document}

\twocolumn[
\icmltitle{Defense Through Diverse Directions}



\icmlsetsymbol{equal}{*}

\begin{icmlauthorlist}
\icmlauthor{Christopher M.~Bender}{to}
\icmlauthor{Yang Li}{to}
\icmlauthor{Yifeng Shi}{to}
\icmlauthor{Michael K.~Reiter}{to}
\icmlauthor{Junier B.~Oliva}{to}
\end{icmlauthorlist}

\icmlaffiliation{to}{The University of North Carolina, North Carolina, USA}

\icmlcorrespondingauthor{Christopher M.~Bender}{bender@cs.unc.edu}

\icmlkeywords{Machine Learning, ICML, diversity, bayesian, adversarial}

\vskip 0.3in
]



\printAffiliationsAndNotice{}  

\begin{abstract}

In this work we develop a novel Bayesian neural network methodology to achieve strong adversarial robustness without the need for online adversarial training.
Unlike previous efforts in this direction, we do not rely solely on the stochasticity of network weights by minimizing the divergence between the learned parameter distribution and a prior.
Instead, we additionally require that the model maintain some expected uncertainty with respect to all input covariates.
We demonstrate that by encouraging the network to distribute evenly across inputs, the network becomes less susceptible to localized, brittle features which imparts a natural robustness to targeted perturbations.
We show empirical robustness on several benchmark datasets.

\end{abstract}

\section{Introduction} \label{sec:introduction}

Neural networks currently achieve greater-than-human performance in a variety of tasks such as object recognition \cite{he2016deep}, language understanding \cite{vaswani2017attention,devlin2018bert}, and game playing \cite{silver2016mastering,silver2017mastering}.
Despite their incredible successes, these same networks are easily fooled by seemingly-trivial perturbations that humans overcome with minimal difficulty \cite{goodfellow2014explaining}.
This weakness poses considerable concern in a world that is increasingly reliant on machines from the perspectives of both security (e.g., face recognition) and safety (driverless cars).
Despite considerable effort to overcome these difficulties, the problem persists \cite{elsayed2018adversarial,athalye2018obfuscated}.

The most successful methods for improving adversarial robustness utilize online adversarial training \cite{madry2017towards}.
Online adversarial training requires an iterative training procedure where adversarial examples are produced based on a particular attack scheme with respect to the current network state and the model is updated to resist the particular attack. 
Unfortunately, this method is computationally expensive, as attacks must be generated and the model updated multiple times.
\citeauthor{zhang2019limitations} and \citeauthor{sharma2017attacking}\ have demonstrated that while this process makes the model robust to the particular type of attack used in the training process, the model can be susceptible to attacks from alternate schemes.

To scale adversarial training, researchers have tried to transfer adversarial examples from another model. \citeauthor{tramer2017ensemble} has found that this offline adversarial training scheme can perform equitably well in practice but can be much more efficient since it decouples the adversarial examples generation from the training process.

Alternative lines of research introduce randomness into the model. Early attempts include adding Gaussian noise to the inputs \cite{zantedeschi2017efficient} and randomly pruning the network \cite{wang2018defensive,dhillon2018stochastic}. \citeauthor{liu2018towards} propose to add Gaussian noise to all the intermediate activations. \citeauthor{wang2019protecting} train multiple copies for each block of the network and randomly select one during inference.
Along these lines, we utilize Bayesian neural networks (BNNs) as a principled way to inject noise into the model.

Recent work \cite{liu2018towards,liu2018adv} has incorporated stochasticity by utilizing BNNs.
Similar to our method, they demonstrate that randomness of BNNs alone is not sufficient for robust classification.
They turn to an online adversarial training scheme to implicitly boost the randomness.

We instead choose to explicitly penalize the model so that it evenly distributes the sensitivity of the output w.r.t.~the input elements.
We estimate the output sensitivity for each input through a first order Taylor approximation and exploit the inherent ensembling of BNNs to evaluate statistics for each input example.
These statistics become the basis for a defense-promoting regularization scheme by diversifying the directions of the output.

Our contributions are as follows:
\begin{itemize}[topsep=0pt]
    \item We propose several general penalties that can be added to the loss function of any Bayesian neural network to diversify the output variation with respect to the input covariates. \\
    \item We demonstrate that increased output diversity leads to natural adversarial robustness, without requiring online adversarial training.
    \item We show that models trained with our diversity inducing penalties generalize to a variety of attack schemes.  
\end{itemize}

This work begins with a review of Bayesian neural networks and the adversarial problem.
We then discuss our motivations and methods for improving model robustness.
Finally, we illustrate our methods on several datasets and discuss the implications.
Particularly, we show that our method improves robustness over state-of-the-art BNN methods \cite{liu2018adv}, all without online adversarial training.

\section{Background \& Related Work} \label{sec:backgroud}

In this section we provide an overview of background material and related work.

\subsection{Bayesian Neural Networks}

In the context of supervised deep learning, a conventional neural network seeks to perform some variant of a classical functional estimation task, to learn a \textit{point estimate} of the optimal function in the chosen functional space that maps each input from the input space to its corresponding output in the output space.
However, such an estimate does not consider, and thus cannot effectively adapt to, the inherent uncertainty throughout the training procedure (e.g.,~data collection, random initialization of network weights).
Such deficiency leads to problems including over-fitting and overly confident predictions. 

To remedy this deficiency, a \textit{Bayesian neural network} (BNN), introduced in the same vein as continuous stochastic processes such as the Gaussian process, seeks to directly model the distribution, whose density we denote as $p$, over random functions
\begin{equation*}
    f \sim p(f),\ \ f: \mathbb{X} \mapsto \mathbb{O}
\end{equation*}
where $\mathbb{X}$ and $\mathbb{O}$ denote the input and the output spaces, respectively.
However, directly learning such a distribution can be arduous as functional spaces are usually infinite-dimensional. Utilizing the fact that neural networks can be regarded as universal approximators for functions \cite{journals/nn/HornikSW89}, a distribution over random functions can be thought of as a choice over of neural networks.
We materialize such connection through learning a distribution over the network weights.
More specifically, in assuming that the network weights are random and distributed according to a prior distribution $P(\textbf{w})$, a BNN seeks to learn the posterior distribution of the network weights, $\textbf{w}$, given the available data, i.e.~$P(\textbf{w}|\textit{D})$.

While a clever idea, learning $P(\textbf{w}|\textit{D})$ is prohibitively expensive even for moderately sized networks because of the high-dimensional integral that needs to be evaluated. Borrowing ideas from variational inference and the recent success of unsupervised methods like the \textit{variational auto-encoder} \cite{DBLP:journals/corr/KingmaW13}, \citeauthor{blundell2015weight} propose to learn a \textit{variational posterior distribution}, $q(\textbf{w}|\theta)$, to approximate the true posterior by optimizing the following objective
\begin{equation}
    \max_{\theta}\ \  \mathbb{E}_{q(\textbf{w}|\theta)}\left(\log P((D|\textbf{w})\right) - \text{KL}\left(q(\textbf{w}|\theta)||p(\textbf{w})\right)
    \label{eq:elbo}
\end{equation}
which is the \textit{evidence lower bound} (ELBO) for the data likelihood.
The expectation term ensures the learnt variational posterior distribution is informed by the data, and the KL divergence acts as a regularizer over the weights.
Although technically any choice of the variational posterior and prior distributions pair is possible, the convention, which we adopt in this work, is to choose both to be independent Gaussians where we learn the mean and variances of the variational posterior distribution.
One benefit of deploying a BNN, which we exploit in our proposed framework, is that for one input one can \textit{draw} multiple functions $f$, which in practice is actualized by drawing a set of different weights from the posterior distribution $P(\textbf{w}|\textit{D})$, to form an ensemble of inferences for various purposes, such as assessing the uncertainties in predictions, gradient evaluations, etc.
In this work, we exploit the network’s variation to control how much sensitivity we expect from each input element.

\subsection{Adversarial Attack} \label{sec:adv_attack}

Adversarial examples are constructed by making small perturbations to the input that induce a dramatic change in the output.
Attacks are typically broken down into two categories: white and black box attacks.
The exact definition of both methods vary, but we will use the following definition in this work.
In the white box setting, the attack has access to the training data set, the fully specified underlying model, and the loss functions.
Attacks are found by performing gradient ascent with respect to the input.  
In the black box setting, the attacker has access to the training data and the loss functions but does not have access to the underlying model parameters.
A black box attack can then be constructed by using a stand-in network trained on the same data with the same loss.
Previous works have shown that these examples are still effective against a variety of other models \cite{liu2016delving,tramer2017space}.

The attacker's goal is:
\begin{equation}
    \max_{\norm{\mathbf{\varepsilon}}_p < \varepsilon_{\textrm{max}}} \E 
        \left[ \mathcal{L}\left( f(\mathbf{x} + \mathbf{\varepsilon}; \parameters), \mathbf{y} \right) \right]
\end{equation}
where $\mathbf{\varepsilon}$ is the attack perturbation, $p$ is the norm (typically taken to be $\infty$), $\varepsilon_{\textrm{max}}$ is the attack budget (the maximum perturbation), $\mathcal{L}$ is the loss, $f$ is the network, $\mathbf{x}$ is the input, $\parameters$ is the network parameters, and $\mathbf{y}$ is the truth.

Typically the attack methods generate the adversarial examples by leveraging the gradient of the loss function with respect to the inputs. The Fast Gradient Sign Method (FGSM) \cite{goodfellow2014explaining}, for instance, takes one step along the gradient direction to perturb an input by the amount $\varepsilon$:
\begin{equation}
    x_{adv} = x + \varepsilon \cdot \text{sign}(\nabla_x \mathcal{L}(f(\mathbf{x}; \parameters), \mathbf{y})).
\end{equation}

Projected Gradient Descent (PGD) \cite{madry2017towards} generalizes FGSM by taking multiple gradient updates:
\begin{equation}
    x_k = x_{k-1} + \alpha \cdot \text{sign}(\nabla_x \mathcal{L}(f(\mathbf{x}; \parameters), \mathbf{y})),
\end{equation}
where $\alpha$ is the step size.
After each update, PGD projects the perturbed inputs back into the $\varepsilon$-ball of the normal inputs.

There are also other types of attacks, such as C\&W attack \cite{carlini2017towards}, Jacobian Saliency Map Attack (JSMA) \cite{papernot2016limitations} and DeepFool \cite{moosavi2016deepfool}. Among all the attack methods, PGD is regarded as the strongest attack in terms of the $L_\infty$ norm. 

\subsection{Adversarial Defense} \label{sec:adv_defense}

The goal of adversarial defense is to render all (bounded) perturbations ineffective against a model.
It is necessary to bound the perturbation or the attack could simply replace an input with an example from a different distribution or class.
A simple intuition to achieve this would be to require that the model be Lipschitz smooth such that
\begin{equation}
    \norm{f(\mathbf{x} + \mathbf{\varepsilon}; \parameters) - f(\mathbf{x}; \parameters)} < C \norm{\mathbf{\varepsilon}}.
\end{equation}
Unfortunately, estimating the Lipschitz coefficient $C$ for an arbitrary network can be extremely difficult, making optimizing over it nontrivial.
\citeauthor{Ciss2017ParsevalNI}\ has attempted to control the Lipschitz coefficient of each layer in the network and, therefore, the network as a whole.

Recent works have introduced a number of defense methods, such as distillation \cite{papernot2016limitations}, label smoothing \cite{hazan2016perturbations}, input denoising \cite{song2017pixeldefend}, feature denoising \cite{xie2019feature}, gradient regularization \cite{ross2018improving}, and preprocessing based approaches \cite{das2017keeping,guo2017countering,buckman2018thermometer}.
Most of these defenses have unfortunately been defeated by subsequent attacks.

The most popular adversarial defense technique incorporates adversarial examples in the training process \cite{madry2017towards}.
Online adversarial training requires generating new examples throughout the training process to map out the local region around known examples and require that the region map to the expected output.
Unfortunately, while this approach does produce a robust defense, it is computationally expensive.

Most similar to our work, ADV-BNN \cite{liu2018adv} attempts to use BNN to combat adversarial attack.
Their proposed method is dependent on online adversarial training and aims to incorporate it into the standard ELBO objective, \refeq{eq:elbo}, as a min-max problem.
In contrast, our proposed methodology does not require online adversarial training, and achieves better performance on CIFAR-10 compared to ADV-BNN.

\subsubsection{Obfuscated Gradients}

\citeauthor{athalye2018obfuscated}\! warn of a failure mode in defense methods that they term ``obfuscated gradients,'' where seemingly high adversarial accuracy is only superficial. 
Networks that achieve apparent improvements in white box attacks through obfuscated gradients do so by making it difficult for an attacker to find $\mathbf{\varepsilon}$.
However, successful $\mathbf{\varepsilon}$ still exist, which indicates that the network has not increased adversarial accuracy despite its improved adversarial test accuracy.
Distinguishing between whether a defense mechanism has increased adversarial accuracy or merely increased attack difficulty is nontrivial.
Typical methods rely on comparing white-box and black-box attacks.
A strong indication that a defense is obfuscating gradients is when black-box attacks are successful while white-box attacks fail (low black-box accuracy and high white-box accuracy).
A defense with high-white box accuracy and fair black-box accuracy is indicative of a mixture of obfuscated gradients and substantive adversarial accuracy improvement.


\section{Motivation} \label{sec:motivation}

Our primary motivation comes from the observation that the fewer inputs an attack needs to perturb, the less robust a model is.
Therefore, we wish to diversify the importance of each input to the output.
Alternatively, we wish to reduce the sensitivity of the output to variations in each input, possibly weighted by some foreknowledge of input uncertainty.

Since adversarial examples are best known for image recognition due to the dramatic difference in human robustness versus machine robustness, we attempt to provide some intuition in this setting.
For the general image setting, the object of interest may exist anywhere in image, and there is no way to know ahead of time which pixels are more reliable.
Therefore, we assume that, on average, the sensitivity due to any single pixel should be roughly equal.
So, we estimate the sensitivity per pixel, normalize across pixels, and penalize any divergence from our expectation.

We can estimate the sensitivity of the $m$th output, $\delta y_m$, with respect to the expected sensitivity of the $d$th input, $\delta x_d$, through a truncated Taylor series
\begin{align}
    \delta y_{m,d}(\mathbf{x}) & \equiv y_m(\mathbf{x} - \delta \mathbf{x}_d) - y_m(\mathbf{x}) \\
    & \approx \delta \mathbf{x}_d^T \frac{\partial y_m(\mathbf{x})}{\partial \mathbf{x}} 
    \ = \ \delta x_d \frac{\partial y_m(\mathbf{x})}{\partial x_d}
\end{align}
and collecting across inputs
\begin{equation}
    \delta \mathbf{y}_m(\mathbf{x}) \approx \delta \mathbf{x} \odot \nabla y_m(\mathbf{x})
\end{equation}
where $\delta \mathbf{y}_m$ is a length $D$ vector corresponding to the sensitivity of the $m$th output with respect to each input.
For simplicity, we assume there is only one output and drop the dependence on $m$.

There are several ways to normalize the result across inputs. We choose to normalize by the $L_2$ norm.
For datasets where each input is equitably important/reliable, $\delta x_d = \delta x$ and the normalized result becomes
\begin{equation} \label{eq:norm_grad}
    \mathbf{u}_y(\mathbf{x}) = \delta \mathbf{y} / ||\delta \mathbf{y}||_2 = 
    \nabla y / \norm{\nabla y}_2
\end{equation}
where $\mathbf{u}_y(\mathbf{x})$ is the direction of the gradient of the the output with respect to the input, $\mathbf{x}$.

This result means that we expect the direction of the output gradient to be uniformly distributed over the unit hypersphere.
In terms of the loss surface, all directions become equally likely.
\begin{equation}
    \mathbf{u}_y \sim U_{\textrm{sph}}(\mathbf{u})
\end{equation}
In the event where the input uncertainty varies, the distribution would be uniformly distributed over a hyper-ellipsoid.

\section{Method} \label{sec:method}

Typically, neural networks are supervised to map an input to an output.
We use the approximation from \refsec{sec:motivation} to further supervise the uncertainty of the output.
However, all networks have some inherent uncertainty (e.g.,~from random initializations) that changes the expected sensitivity.
We execute $K$ draws of the BNN and include additional penalties that attempt to maintain the expected distribution of the output sensitivity.
We experiment with a variety of penalties based on this premise.
For the sake of brevity, we drop the dependence of  $\textbf{u}_y$ on $\textbf{x}$.

\subsection{Entropy and Variances}

As motivated previously, in order to increase the network's robustness against adversarially perturbed input, we encourage the network to maintain, and hence to evenly distribute, some expected sensitivity with respect to all input covariates.
We materialize this idea by encouraging the normalized gradient in \refeq{eq:norm_grad} to be uniformly distributed over the unit hypersphere, which in turn is equivalent to maximizing the entropy of $\textbf{u}_y$,\ denoted as $H\!\left(\textbf{u}_y \right)$, because $\textbf{u}_y$ is bounded within the unit hypersphere.
However, as a function of the network weights $\textbf{w}$, the density of $\textbf{u}_y$ is intractable even for moderate-sized network, making maximizing $H\!\left(\textbf{u}_y \right)$ directly prohibitively expensive and impractical in practice. 

In a simplified setting where the elements are independent, maximizing the sum of the variances of each element is equivalent to maximizing the entropy of the random vector.

\begin{prop}
    Given a random vector $\textbf{X} = \left(X_1, X_2, \cdots, X_D\right)^T$ where its elements are independent and $X_i \in [a_i,b_i]$ for all $i$, there exists a monotonically increasing relationship between the entropy of \textbf{X}, $H(\textbf{X})$, and the sum of the variances of each element of $\textbf{X}$, $\sum_i \var(X_i)$. 
\end{prop}

Proposition 1 indicates that maximizing the entropy of $\textbf{X}$ is equivalent to maximizing $\sum_i\text{Var}(X_i)$.
See Appendix A for the proof.
While $\textbf{u}_y$ is not independent in our case, we use Prop.~1 as an analogy and adopt the sum of the variances of the elements of $\textbf{u}_y$ as a surrogate for $H\!\left(\textbf{u}_y \right)$.

\subsection{\directloss}


A simple method might be to maximize the sum over inputs of the variance of $\mathbf{u}_y$ across the $K$ draws
\begin{equation}
    \sum_{d=1}^D \var \left[ u_{y,d} \right].
\end{equation}
However, since $\mathbf{u}_y$ is a unit vector, this loss degenerates and only serves to minimize the average value of the output sensitivity
\begin{align}
    \sum_{d=1}^D \var \left[ u_{y,d} \right] & = 
        \E \left[ \sum_{d=1}^D u_{y,d}^2 \right] - \sum_{d=1}^D \E \left[ u_{y,d} \right]^2 \\
    & = 1 - \sum_{d=1}^D \E \left[ u_{y,d} \right]^2.
\end{align}

Since we wish to maximize the variances w.r.t. $\mathbf{u}_{y}(\mathbf{x})$ we consider the mean penalty $\Omega_M(x)$:
\begin{align}
    \Omega_M(\mathbf{x}) = \sum_{d=1}^D \E \left[ u_{y,d} \right]^2.
    \label{eq:meanpen}
\end{align}

\subsection{\minvar}

While Equation \eqref{eq:meanpen} increases the total variance across inputs, it does not necessarily encourage diversity.
We consider several additional penalties to include with \refeq{eq:meanpen} that do increase diversity.


A simple penalty to increase the minimum variance over dimensions is
\begin{align} \label{eq:minvarloss}
    \Omega_{V,1}(\mathbf{x}) = -\min_{d} \var \left[ u_{y,d} \right].
\end{align}
Equation \eqref{eq:minvarloss} exploits the fact that the sum (over dimensions) of the variance is fixed.
Therefore, increasing the minimum necessarily decreases the other values.
In theory, as the network trains, the minimum element changes and eventually all the elements converge to the same variance.

Unfortunately, since the loss only supervises one pixel at a time, the minimum shifts across a few elements and never influences the pixels with the largest variance.
We consider two simple methods to correct this difficulty.
One is to supervise all the pixels simultaneously by replacing the $\min$ operation with a soft-min weighted sum so that \refeq{eq:minvarloss} becomes
\begin{equation} \label{eq:var_sm}
    \Omega_{V,2}(\mathbf{x}) = -\sum_{d=1}^D \textrm{softmin}\!\left( \alpha u_{y,d} \right) \cdot \var \left[ u_{y,d} \right]
\end{equation}
where $\alpha$ corresponds to the temperature.
However, since the sum of the variances is fixed and this loss attempts to increase the variance of all pixels simultaneously, we find that it can result in a counterproductive competition across pixels.

A more direct method to supervise the variance is to minimize the distance between the observed variance and the expected variance of $1/N$.
Using the Euclidean distance, \refeq{eq:minvarloss} becomes
\begin{equation} \label{eq:var_f}
    \Omega_{V,3}(\mathbf{x}) = \sum_{d=1}^D \left( \var \left[ u_{y,d} \right] - 1/D \right)^2.
\end{equation}
We find that this final representation yields the best results amongst the variance encouraging losses and choose it as the variance penalty $\Omega_V(\mathbf{x})$.

\subsection{Non-Sparse Promoting Losses} \label{sec:dpl}

Directly encouraging the model to match a specific distribution's second order moment may be too strict a requirement.
As an alternative to matching the second moment of the uniform hypersphere, we consider penalizing \textit{small} $L_1$ norms of $\mathbf{u}_y$.
By requiring that the $L_1$ norm be large, we bias the network away from over reliance on a few features and encourage greater diversity across the input covariates.
Unlike in the variance-based losses, the network does not have to match each input direction to the same value.
This allows for increased flexibility while still maintaining the same intuitive effect on the diversity of dependence.
The added penalty becomes
\begin{equation} \label{eq:dpl}
    \Omega_S(x) = -\E \left[ \norm{\mathbf{u}_y}_1 \right].
\end{equation}

\subsection{Batch Loss}
We summarize the full possible loss of our model with respect to posterior parameters, $\mathcal{L}(\parameters)$, with all the above penalties, a supervised loss $l$ and a batch of data $\{ (x_n, y_n) \}_{n=1}^N$:
\begin{equation} \label{eq:full_loss}
    \begin{split}
        \mathcal{L}(\parameters) = 
        \frac{1}{N} \sum_{n=1}^N \Big{[} l(x_n, y_n) + \lambda_M \Omega_M(x_n) \ \ \ \ \ \ \  \\
        \ \ \ + \lambda_V \Omega_V(x_n) + \lambda_S \Omega_S(x_n) \Big{]}.
    \end{split}
\end{equation}
We vary \refeq{eq:full_loss} below, by setting various penalty weights $\lambda$ to zero.

\subsection{Further Benefits of Adversarial Training} \label{sec:adt}

Finally, we test the benefits of offline adversarial training \cite{tramer2017ensemble} in addition to the diversity inducing penalties.
Adversarial examples are pre-computed from a trained, non-Bayesian neural network of a matching architecture and then added statically to the training set.
Thus, this form of defense is more efficient than online adversarial training, which requires on-the-fly computation of adversarial examples.

\section{Experiments} \label{sec:exp}

In this section we explore the effect of inducing diversity on a variety of open-source datasets.

\subsection{Penalty Shorthand}

We present our results by adding different combinations of the diversity-encouraging penalties in \refsec{sec:method}.
To declutter the results, we use the following shorthand to refer to the different penalties we apply to the model.
The unit gradient mean penalty, \refeq{eq:meanpen}, is given as ``M;'' the variance penalty, \refeq{eq:var_f}, by ``V;'' the non-sparse penalty, \refeq{eq:dpl}, by ``S;'' and any offline training by ``Off.''
We additionally denote when a network is Bayesian by prepending the network name with a ``B.''
For example, the case where we use a Bayesian VGG16 with the mean and variance penalty is shorthanded as ``BVGG16-M-V.''

As mentioned previously, all adversarial examples appended to the training set for offline training were constructed using conventional networks with matching architectures.
Aside from the models trained with offline adversaries, we do not include any form of data augmentation.

\subsection{Practical Considerations} \label{sec:arch}

In this section we discuss several practical steps taken to implement various networks and diversity-promoting penalties.
We utilize TensorFlow \cite{tensorflow2015-whitepaper} and Tensorflow Probability \cite{DBLP:journals/corr/abs-1711-10604} to implement the general and probabilistic components of our models, respectively.

\subsubsection{Drawing from the Bayesian Network}

Since we optimize over statistics of the probabilistic network, we require multiple network samples for the same input data.
A simple method to exploit extant parallelisms in most neural network frameworks would be to duplicate each element in the batch $K$ (the number of draws) times.
Unfortunately, Bayesian networks implement the Bayesian layers using either FlipOut \cite{DBLP:journals/corr/abs-1803-04386} or the reparameterization trick \cite{DBLP:journals/corr/KingmaW13} to efficiently draw parameters that are \textit{shared} across each element in the batch.
While this shortcut is sufficient to decorrelate training gradients, it is not sufficient to obtain the level of independence required to inform our methods.
As a result, we resort to executing the network $K$ times for \textit{each batch}.

This becomes prohibitively expensive for networks of sufficient depth.
To offset some of this cost, we break our networks into two parts.
In the first part, the layers are constructed in the conventional fashion without Bayesian components.
We will refer to this component of our models as the deterministic network.
After the deterministic network, we construct a Bayesian network.
The deterministic component is executed once per batch and the Bayesian component, $K$ times.
The gradient that informs diversity promotion is still taken with respect to the input of the full network.
This means that while the deterministic component does not directly add variation it is still trained to encourage overall network diversity.
The specifics of where the transition between determinism and Bayesian can be found in each experiment's section.
In general, we found that it was sufficient to set the last quarter of the network as Bayesian and use $K=10$ draws.

\subsubsection{Attack Schemes} \label{sec:attack_types}

We test our defense against two types of common attacks: the Fast-Gradient Sign Method (FGSM) and Projected Gradient Descent (PGD).
We deploy both $L_\infty$ and $L_2$ attacks.  
Black box attacks are performed using examples from external sources when available.
All attacks are performed using the Adversarial Robustness Toolbox \cite{art2018} and use the same number of network draws as are used in training.

\subsection{MNIST} \label{sec:mnist}

We test our diversity induced models on MNIST \cite{lecun-mnisthandwrittendigit-2010}.
For these tests we use a simple CNN with three convolutional layers followed by two fully connected layers.
Since this network is fairly shallow, we compose the deterministic part of the network using the first two convolutional layers and use Bayesian layers for the final convolutional and both fully connected layers.
When the corresponding penalty is used, the loss hyperparameters are: $\lambda_M=20$, $\lambda_V=40$, and $\lambda_S=40$.

\begin{table}[t]
    \centering
    \caption{Adversarial accuracy (\%) on MNIST with various combinations of diversity promotion against $L_\infty$ attacks.}
    \label{tab:mnist_acc_linf}
    \begin{tabular}{c c c c c c}
        \toprule
        Method    &     FGSM  &      PGD  & Black-Box \\
        \midrule
        CNN       &     36.8  &      2.6  &     57.8  \\
        BNN       &     55.2  &      0.9  &     56.4  \\
        BNN-Off   &     40.0  &      2.1  & \bf{93.6} \\
        M         &     93.0  &     55.2  &     60.1  \\
        M-V       &     94.0  &     70.0  &     61.8  \\
        M-S       &     96.3  &     94.0  &     54.3  \\
        M-V-S     &     97.2  &     95.8  &     59.8  \\
        M-S-Off   & \bf{97.6} & \bf{95.8} &     89.9  \\
        M-V-S-Off &     97.4  &     94.5  &     90.6  \\
        \bottomrule
    \end{tabular}
\end{table}

Table \ref{tab:mnist_acc_linf} shows the accuracy of the model when trained using different combinations of diversity inducing penalties and adversarial training against $L_\infty$ attacks with a maximum attack budget of 0.3.
All models obtain better than 99\% standard accuracy.
Black box attacks were obtained using examples from an open repository.\footnote{https://github.com/MadryLab/mnist\_challenge}

When only the mean penalty (\refeq{eq:meanpen}) is imposed, the model shows modest improvements in both forms of attack; indicating that the defense has succeeded in improving the robustness of the model against attack.
While this defense has not completely succeeded in overcoming attacks, it is a positive step.
As we include additional penalties, we observe that the white-box attack improves, most notably with the inclusion of the non-sparse penalty (\refsec{sec:dpl}).
However, these models generally show a slight \textit{decrease} in black-box performance.
We infer that these penalties tend to cause the model to over fit the defense by obfuscating gradients.
Fortunately, offline adversarial training appears to sufficiently augment the training set so that the defenses can generalize.

\subsubsection{MNIST Accuracy Evolution}

\begin{figure}[ht!]
    \centering
    \begin{subfigure}{0.48\textwidth}
        \centering
        \includegraphics[width=\textwidth]{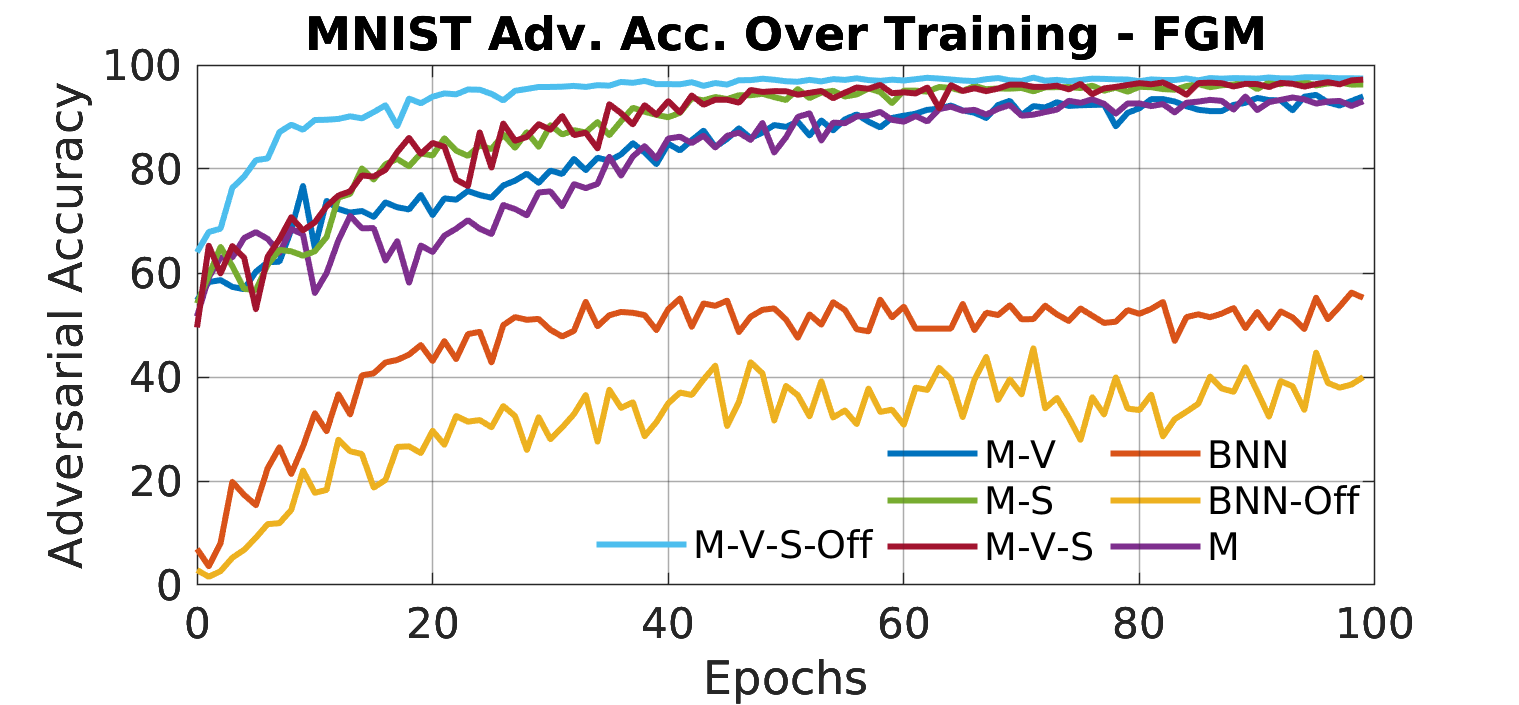} \\
        \caption{FGSM $L_\infty$ $\varepsilon=0.3$ \\ \ \\}
    \label{fig:mnist_evolve_wacc_fgsm}
    \end{subfigure}
    
    \begin{subfigure}{0.48\textwidth}
        \centering
        \includegraphics[width=\textwidth]{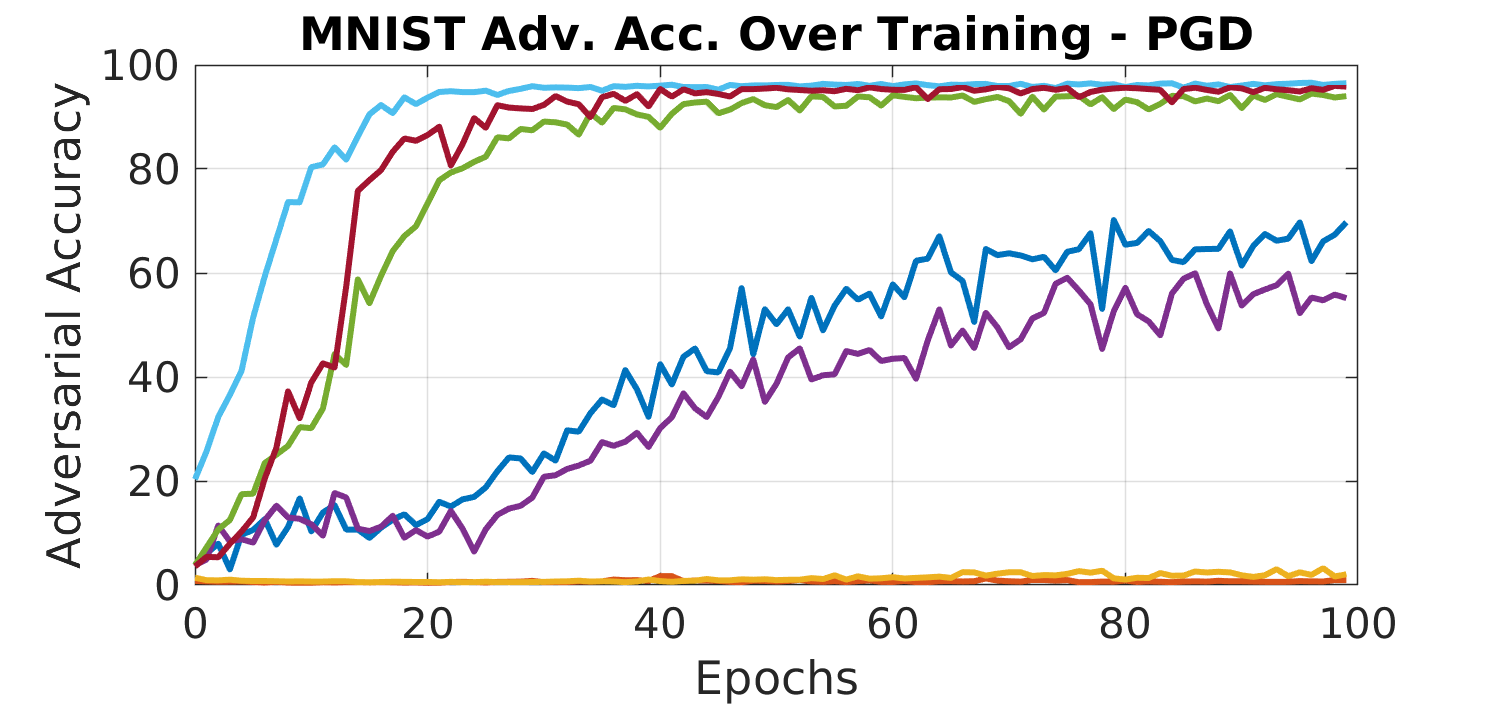} \\
        \caption{PGD $L_\infty$ $\varepsilon=0.3$}
    \label{fig:mnist_evolve_wacc_pgd}
    \end{subfigure}
    
    \caption{Evolving white box attack accuracy after each epoch.}
    \label{fig:mnist_evolve_wacc}
\end{figure}

Figure \ref{fig:mnist_evolve_wacc} illustrates how the adversarial accuracy evolves over the training process.
We chose to train for 100 epochs to give the defenses adequate opportunity to influence the model.
Figure \ref{fig:mnist_evolve_wacc_fgsm} shows attack accuracy against the FGSM and  \reffig{fig:mnist_evolve_wacc_pgd} shows accuracy against PGD.
In both cases, attacks are generated using the \textit{test} set against the current model state.
These figures provide some insight into how the defense is instilled in the model as it is trained.
The mean and variances penalties appear to slowly (but consistently) influence the model throughout the training process.
These penalties seem to still be improving the adversarial accuracy against PGD attacks even after 100 epochs have elapsed, well after the standard accuracy has converged.

As speculated in \refsec{sec:dpl}, the non-sparse penalty appears to give the model greater flexibility and is easier to learn.
This quick increase and the disparity between white and black box performance in Table \ref{tab:mnist_acc_linf} may indicate that the penalty is prone to causing the defense to over fit by obfuscating gradients.
However, this is assuaged with the use of offline adversarial training.
All models converged to a standard accuracy of 99\% within the first two epochs.

\subsection{CIFAR-10} \label{sec:cifar10}

In this section, we evaluate our proposed diversity inducing penalties on the CIFAR-10 dataset \cite{cifar10_citation}.
The backbone network is VGG16.
A Bayesian version of VGG16 is built by replacing the last convolutional block and the fully connected layers with variational alternatives.
When training with our proposed penalties, we use the hyparameters: $\lambda_M = 10$ and $\lambda_S=20$.
We report the $L_\infty$ attack with a budget of $\varepsilon = 0.03$ here.
PGD attack perform 40 gradient updates with a step size 0.001.
Refer to Sec.~\ref{sec:ablation} for ablation experiments on the attack budget.
We report the black box attack accuracy using examples from an open repository.\footnote{https://github.com/MadryLab/cifar10\_challenge}

Table \ref{tab:c10_acc} demonstrates the accuracy of the defended models on CIFAR-10.
The standard loss is included in the table since it varies across models.
Given the marginal improvements in adversarial accuracy from the variance-based penalties, we forego their use in this case.

Unlike in the MNIST experiments, the mean penalty is insufficient to overcome white box attacks; however, it does offer improvements in the black box setting.
Also unlike MNIST, these results do not indicate that our methods suffer from the obfuscated gradients problem as white box attacks are consistently more effective than black box attacks.
The additional data augmentation from offline adversarial training further improves defense.

\begin{table}[]
    \centering
    \small
    \caption{Adversarial accuracy (\%) under $L_\infty$ white box attack and black box attack on CIFAR-10 with various combinations of diversity promotion.}
    \label{tab:c10_acc}
    \begin{tabular}{c c c c c}
        \toprule
        Loss       &      Std. &      FGSM &       PGD & Black-Box \\
        \midrule
        VGG16      & \bf{90.4} &     14.4  &      5.9  &     58.0  \\
        BVGG16     &     89.1  &     12.9  &      5.6  &     24.5  \\
        BVGG16-Off &     85.8  &     58.8  &     57.0  &     84.3  \\
        M          &     90.1  &     14.2  &      6.1  &     64.1  \\
        M-S        &     88.4  &     50.2  &     47.7  &     55.2  \\
        M-S-Off    &     86.3  & \bf{73.1} & \bf{72.3} & \bf{84.8} \\
        \bottomrule
    \end{tabular}
\end{table}
\normalsize

Table \ref{tab:c10_advbnn_comp} juxtaposes several of our proposed defenses against results reported in \citeauthor{liu2018adv} for several attack budgets.
We observe that our method without any adversarial training maintains higher standard accuracy and shows improved robustness to higher attacks budgets.
Surprisingly, we observe that a Bayesian VGG16 with offline adversarial training obtains consistent accuracy above the other methods.
We suspect this is because the adversarial examples used in offline training were constructed using PGD with 40 steps whereas the examples in ADV-BNN were defended using online training with 10 steps.
Including our penalties consistently increases the performance of the offline model by approximately 15\%.

\begin{table}[]
    \centering
    \caption{Comparison of accuracy (\%) under PGD attack with different budget. We use the same attack as in \cite{liu2018adv} to make comparisons fair. Results for ``Adv-CNN'' and ``Adv-BNN,'' representing adversarially trained CNN and BNN, are from \cite{liu2018adv}. Note, we use offline generated adversarial examples to train our model, which is much more efficient.}
    \label{tab:c10_advbnn_comp}
    \begin{tabular}{cccccc}
    \toprule
    Method/Budget  & 0.0       & 0.015      & 0.035      & 0.055      & 0.07 \\
    \midrule
        Adv-CNN    & 80.3      & 58.3       & 31.1       & 15.5       & 10.3 \\
        Adv-BNN    & 79.7      & 68.7       & 45.4       & 26.9       & 18.6 \\
        M-S        & \bf{88.4} & 61.3       & 47.8       & 39.8       & 34.3 \\
        BVGG16-Off & 85.8      & 58.2       & 57.8       & 57.3       & 57.3 \\
        M-S-Off    & 86.3      & \bf{73.2}  & \bf{73.1}  & \bf{73.0}  & \bf{73.0} \\
    \bottomrule
    \end{tabular}
\end{table}

\section{Ablation}\label{sec:ablation}

\begin{figure}[t!]
    \centering
    
    \begin{subfigure}{0.48\textwidth}
        \centering
        \includegraphics[width=\textwidth]{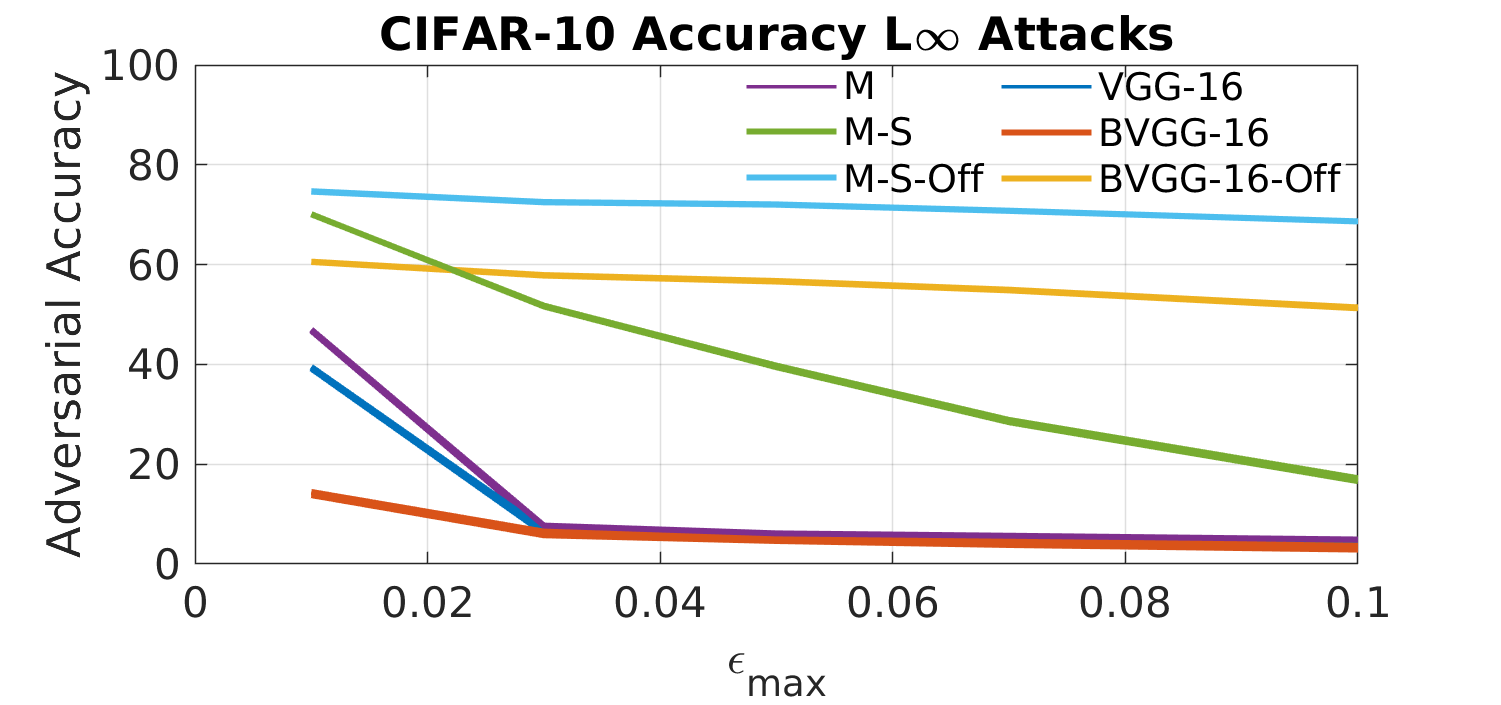}
        \caption{$L_\infty$ attack budgets.}
        \label{fig:cifar10_linf_budgets}
    \end{subfigure}
    
    \begin{subfigure}{0.48\textwidth}
        \centering
        \includegraphics[width=\textwidth]{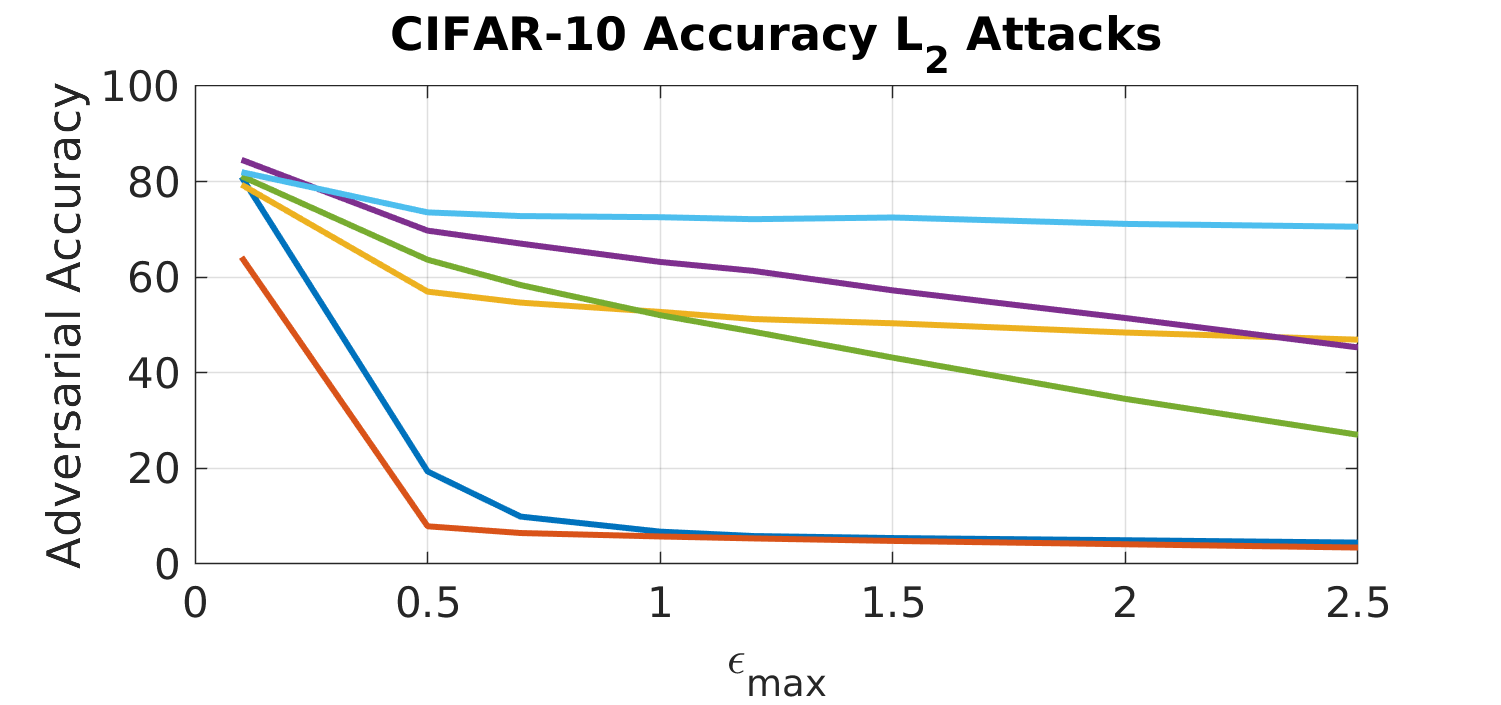}
        \caption{$L_2$ attack budgets.}
        \label{fig:cifar10_l2_budgets}
    \end{subfigure}
    
    \caption{Varying attack budgets with 40 step PGD attacks against CIFAR-10 defenses.}
    \label{fig:cifar_ablation}
\end{figure}

To test our defenses' efficacy under more diverse attack cases, we vary the attack budget with PGD against our models trained on CIFAR-10.
Figure \ref{fig:cifar_ablation} illustrates the results from all of our tests.

Figure \ref{fig:cifar10_linf_budgets} demonstrates how the various models respond to increases in the attack budget of PGD $L_\infty$ attacks.
Unsurprisingly, all the models show decreases in performance as the attack budget is increased.
The best model is consistently the mean and non-sparse penalized models with offline adversarial training.
The diversity penalties consistently improve the accuracy over offline training by approximately 15\%.
Strangely, the mean and non-sparse encouraging penalties without offline augmentation experience a steady decrease in performance and is quickly outpaced by offline augmentation.

Figure \ref{fig:cifar10_l2_budgets} uses $L_2$ attacks but otherwise demonstrates the same variation as in \reffig{fig:cifar10_linf_budgets} (changes in the attack budget of PGD).
The most interesting feature is how well the mean penalty performs without any additional augmentations.
It begins with the highest accuracy and is consistently higher than the model trained with the mean and non-sparse penalties and is primarily outperformed only by the full mean and non-sparse penalized models with offline training.
Otherwise, the results are consistent with those found in the $L_\infty$ study.

\section{Discussion} \label{sec:discussion}

We have demonstrated the efficacy of explicitly encouraging diversity of the output with respect to the input.
On MNIST, we show that we can obtain strong adversarial robustness without the need for any form of adversarial training.
In this case, our black box accuracy falls short of the white box accuracy, indicating we may be obfuscating gradients.
Fortunately, the black box performance is still fair, which may mean that our method improves model robustness and obfuscates gradients.
Including offline adversarial training in these models improves the black box accuracy so that it is on par with the white box accuracy.
We suspect that the original MNIST training set is not diverse enough itself and that additional data or augmentations may be sufficient to prevent the defense from over fitting and obfuscating gradients.

Our results on CIFAR-10 are quite encouraging.
We demonstrate that our method is capable of improving adversarial accuracy with only a small reduction in standard accuracy.
These models do not appear to suffer from the obfuscated gradients problem: black box accuracy is consistently higher than white box performance.
Further, the ablation studies show a consistent reduction in accuracy as attack strength is increased.
Finally, our model compares favorably with other Bayesian defense mechanisms, achieving superior performance in most cases without any adversarial training.
The added use of offline adversarial training improves our models' performance so that they are superior in all cases.

We speculate that it may be possible to further improve our defense results by including additional forms of standard augmentation, e.g.~the addition of Gaussian noise, shifting, etc.
Similarly, we performed only a small search for the hyperparameters of the diversity penalties, $\lambda$, and used them for all penalty combinations and regardless of the use of offline training.
Additional gains may be possible by performing a finer-grained search over these parameters.

\section{Conclusions} \label{sec:conclusions}

In this work, we built upon the Bayesian neural network framework and introduced a novel technique, namely Defense through Diverse Directions, to achieve strong adversarial robustness.
This is a daunting task, where models that utilize the concept of randomness to combat adversarial attack were previously limited to adding random noise layers to the network \cite{liu2018towards} or simply applying a Bayesian neural network to take advantage of the stochasticity of the weights \cite{liu2018adv}.
To the best of our knowledge, this is the first attempt to attain adversarial robustness through explicitly requiring the network to maintain and evenly distribute expected and sensible uncertainty with respect to input covariates, achieved by the various penalty terms we introduce.
Without the need for online adversarial training, we demonstrate the effectiveness and robustness of our approach by achieving the strong adversarial (after-attack) accuracies on various datasets against different adversarial attacks.

\newpage



\bibliography{references.bib}
\bibliographystyle{icml2020}

\newpage
\appendix

\section{Proof of Proposition 1}
\begin{proof}
	Suppose $\textbf{X} \sim f_\textbf{X}(\textbf{x})$. Since the elements of $\textbf{X}$ are independent, the entropy of the random vector \textbf{X} equals the sum of the entropy of each individual element
	
	\begin{equation}
		\begin{split}
		H(\textbf{X}) & = -\int_{\textbf{x}} f(\textbf{x})\ \log f(\textbf{x}) d\textbf{x} \\
		 & = - \sum_{i}\int_{\textbf{x}} \left[\prod_j f(x_j)\right] \log f(x_i) d\textbf{x}\\
		 & = - \sum_{i}\int_{x_i}  f(x_i)\log f(x_i) \ dx_i\\
		 & = \sum_i H(X_i)
		 \end{split}
	\end{equation}

    and that 

	\begin{equation*}
		\begin{split}
		H(X_i) & = \int_{x_i}  f(x_i)\log \frac{1}{f(x_i)} \ dx_i \\
		 & =  \int_{x_i}  (x_i-\mu_i)^2 f(x_i) \log \left(e^{\frac{1}{(x_i-\mu_i)^2}} + \frac{1}{f(x_i)}\right)\ dx_i. \\
		 \end{split}
	\end{equation*}
	
Since by assumption each $X_i$ is bounded in a compact interval, we have $\log \left(e^{1/(x_i-\mu_i)^2} + 1/f(x_i)\right) > 1$. It also achieves its maximum and minimum values on this compact interval which we denote as $k$ and $h$, respectively. We then have
$$h \cdot \var[X_i] \leq H(X_i) \leq k\cdot \var[X_i]$$

indicating that maximizing the entropy of $\textbf{X}$ is equivalent to maximizing $\sum_i\text{Var}(X_i)$. 
\end{proof}

\vfill\eject

\section{Additional Results}

\begin{table}[h!]
    \centering
    \begin{tabular}{c c c c c c}
        \toprule
        Method    & BIM & CW & FGM & PGD \\
        \midrule 
        CNN       &  7.1 & 30.5 & 36.8 &  2.6 \\
        BNN       &  5.5 & 88.9 & 55.2 &  0.9 \\
        BNN-Off   &  4.8 &   -  & 40.0 &  2.1 \\
        M         & 88.2 & 73.0 & 93.0 & 55.2 \\
        M-V       & 89.5 & 67.4 & 94.0 & 70.0  \\
        M-S       & 97.1 & 75.4 & 96.3 & 94.0 \\
        M-V-S     & 97.8 & 78.8 & 97.2 & 95.8 \\
        M-S-Off   & 97.2 & 65.6 & 97.6 & 95.8 \\
        M-V-S-Off & 97.5 & 61.9 & 97.4 & 94.5 \\
        \bottomrule
    \end{tabular}
    \caption{Adversarial accuracy on MNIST with various combinations of diversity promotion against $L_\infty$ attacks.}
    \label{tab:mnist_acc_linf_full}
\end{table}

\begin{figure}[ht!]
    \centering
    \begin{subfigure}{0.5\textwidth}
        \centering
        \includegraphics[width=\textwidth]{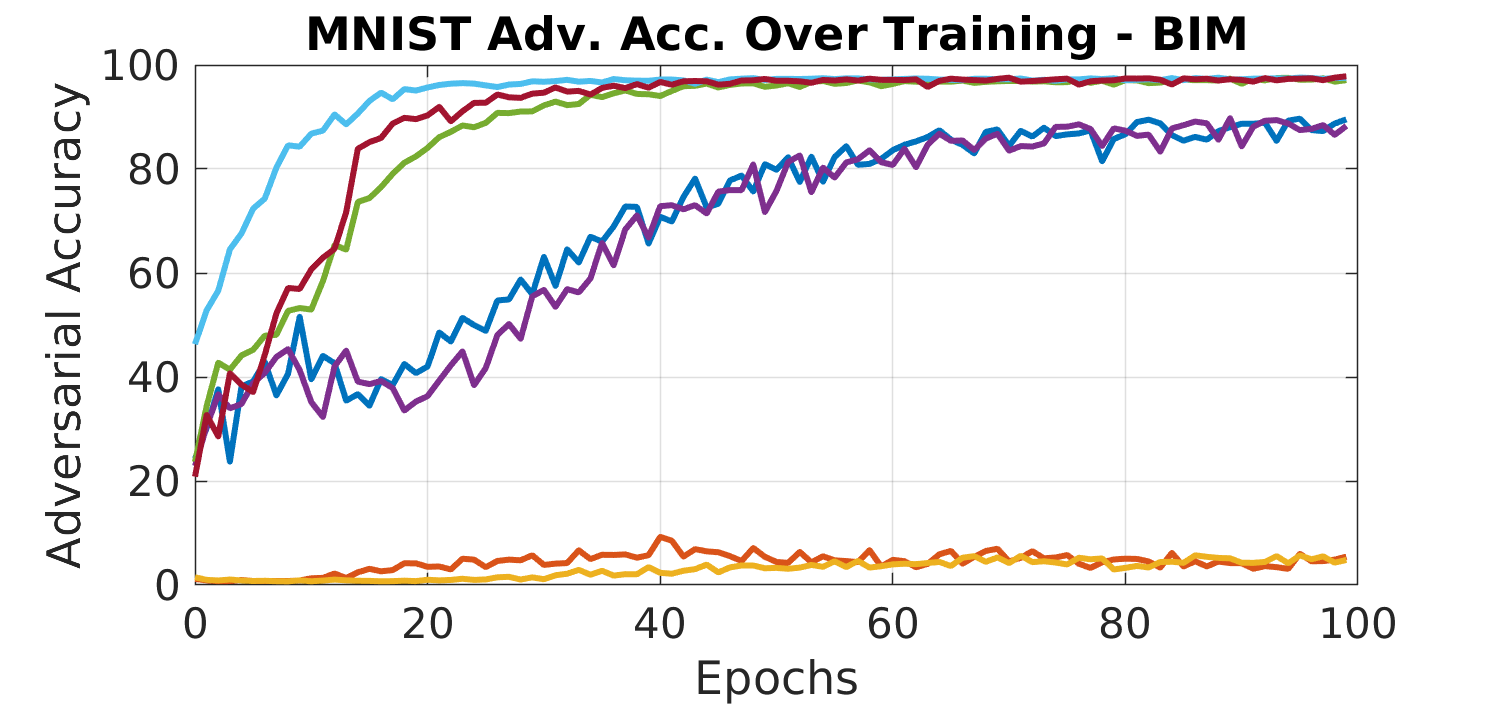} \\
        \caption{art attack}
    \end{subfigure}
    
    \begin{subfigure}{0.5\textwidth}
        \centering
        \includegraphics[width=\textwidth]{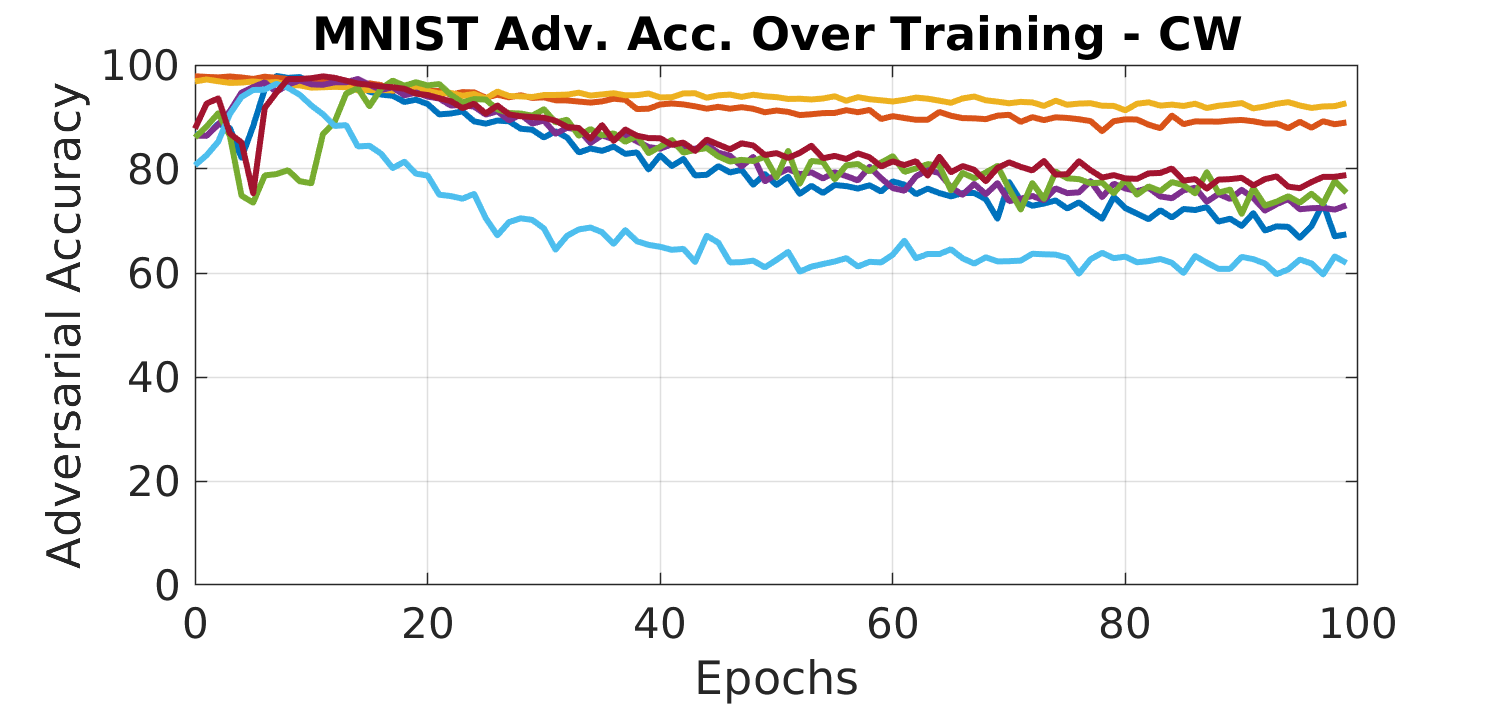} \\
        \caption{art attack}
    \end{subfigure}
    
    \caption{White box attack accuracy after each epoch.}
    \label{fig:art_attack}
\end{figure}


\end{document}